\def\year{2021}\relax

\documentclass[letterpaper]{article} 
\usepackage{aaai21}  
\usepackage{times}  
\usepackage{helvet} 
\usepackage{courier}  
\usepackage[hyphens]{url}  
\usepackage{graphicx} 
\usepackage{natbib}  
\usepackage{caption} 
\usepackage{bm}
\usepackage{amsmath}
\usepackage{booktabs}

\nocopyright
\title{Skip-Connected Self-Recurrent Spiking Neural Networks with \\ Joint Intrinsic Parameter and Synaptic Weight Training}
\author {
    Wenrui Zhang
        \textsuperscript{\rm 1},
        	Peng Li
        \textsuperscript{\rm 2} \\
}
\affiliations {
    University of California, Santa Barbara\\ Santa Barbara, CA 93106 \\
   \textsuperscript{\rm 1}wenruizhang@ucsb.edu,\textsuperscript{\rm 2}lip@ucsb.edu\\
}

\begin{document}

\maketitle

\begin{abstract}
As an important class of spiking neural networks (SNNs), recurrent spiking neural networks (RSNNs) possess great computational power and have been widely used for processing sequential data like audio and text. However, most RSNNs suffer from two problems. 1. Due to a lack of architectural guidance, random recurrent connectivity is often adopted, which does not guarantee good performance. 2. Training of RSNNs is in general challenging, bottlenecking achievable model accuracy. To address these problems, we propose a new type of RSNNs called Skip-Connected Self-Recurrent SNNs (ScSr-SNNs). Recurrence in ScSr-SNNs is introduced in a stereotyped manner by adding self-recurrent connections to spiking neurons, which implements local memory. The network dynamics is enriched by skip connections between nonadjacent layers. Constructed by simplified self-recurrent and skip connections, ScSr-SNNs are able to realize recurrent behaviors similar to those of more complex RSNNs while the error gradients can be more straightforwardly calculated due to the mostly feedforward nature of the network. Moreover, we propose a new backpropagation (BP) method called backpropagated intrinsic plasticity (BIP) to further boost the performance of ScSr-SNNs by training intrinsic model parameters. Unlike standard intrinsic plasticity rules that adjust the neuron's intrinsic parameters according to neuronal activity, the proposed BIP methods optimizes intrinsic parameters based on the backpropagated error gradient of a well-defined global loss function in addition to synaptic weight training. Based upon challenging speech and neuromorphic speech datasets including TI46-Alpha, TI46-Digits, and N-TIDIGITS, the proposed ScSr-SNNs can boost performance by up to $2.55\%$ compared with other types of RSNNs trained by state-of-the-art BP methods.
\end{abstract}

\section{Introduction}
Recurrent neural networks (RNNs) are one of the most popular types of networks in artificial neural networks (ANNs). They are designed to better handle sequential information such as audio or text. RNNs make use of internal states to store past information, which is combined with the current input to determine the current network output. During the past few decades, RNNs have been widely studied with various structures such as Gated Recurrent Units (GRU)~\cite{cho2014learning}, Long Short Term Memory (LSTM)~\cite{hochreiter1997long}, Legendre Memory Units (LMU)~\cite{voelker2019legendre}, Echo state networks (ESN)~\cite{jaeger2001echo}, and Deep RNNs~\cite{graves2013speech}.

Similarly, recurrent spiking neural networks (RSNNs) are competent for processing temporal signals such as time series or speech data~\cite{ghani2008neuro}. On the other hand, RSNNs are advantageous from a biological plausibility point of view compared with their non-spiking counterparts and enjoy their intrinsic spatiotemporal computing power attributed to the rich network dynamics created by the recurrence of connectivity.  While mature network architectures have been developed for  RNNs, the complex dynamics of RSNNs are not well understood and training of RSNNs is in general challenging. These difficulties lead to the demonstration of simple RSNN architectures, severely limiting the practical application of RSNNs. 

One of the well-known RSNN models is the liquid State Machine (LSM)~\cite{maass2002real} which has a single randomly connected recurrent reservoir layer followed by one readout layer. The reservoir weights typically are either fixed or trained by unsupervised learning like spike-timing-dependent plasticity  (STDP)~\cite{morrison2008phenomenological} with the readout layer only trained by supervision~\cite{ponulak2010supervised,zhang2015digital,jin2016ap}. In recent years, the reservoir of LSM has been extended to different structures. In \cite{wang2016d} and \cite{srinivasan2018spilinc}, multiple reservoirs are applied to process different parts of the input signals. \cite{bellec2018long} proposed an architecture called long short-term memory SNNs which can be considered as a structured reservoir. It has a regular spiking portion with both inhibitory and excitatory spiking neurons and an adaptive neural population. Like in typical LSM models, the recurrent connections between neurons are sparsely and randomly generated with a certain probability. \cite{zhang2019spike} demonstrates training of deep RSNNs by a backpropagation method called ST-RSBP. However, recurrent connections in these RSNNs are still randomly generated. 

On the other hand, \cite{shrestha2017spike} and \cite{lotfi2020long} propose to convert a pre-trained ANN-based LSTM to a spiking LSTM. However, they either directly substitute the original non-spiking activation function with a spiking activation function or approximate the non-spiking activation function heuristically. The original ANN LSTM model is re-trained using a standard BP method for ANNs and no additional training is applied to the converted SNN LSTM model in the spatiotemporal domain. Hence, these ANN-to-SNN conversion approaches are unable to explore the intrinsic spatiotemporal computing capability of typical spiking neurons such as ones modeled using the Leaky-Integrate-and-Fire (LIF) model or spike response model (SRM).


In this paper, we propose a new  RSNN model called Skip-Connected Self-Recurrent SNNs (ScSr-SNNs) to offer a simple and structured approach for designing high-performance RSNNs and mitigating the training challenges resulted from random recurrent connections as in the prior work.  In each recurrent layer of an ScSr-SNN, recurrence is only introduced by self-recurrent connections of individual spiking neurons, i.e., there exist no lateral connections between different neurons, while the self-recurrent connections also implements local memory. The network dynamics is enriched by skip connections between nonadjacent layers. Constructed by simplified self-recurrent and skip connections, ScSr-SNNs are able to realize recurrent behaviors similar to those of more complex RSNNs while the error gradients can be more straightforwardly calculated due to the mostly feedforward nature of the network. Moreover, we propose a new backpropagation (BP) method called backpropagated intrinsic plasticity (BIP) to further boost the performance of ScSr-SNNs by training intrinsic model parameters. Unlike standard intrinsic plasticity rules that adjust the neuron's intrinsic parameters according to neuronal activity, the proposed BIP method optimizes intrinsic parameters based on the backpropagated error gradient of a well-defined global loss function in addition to synaptic weight training. Based upon challenging speech and neuromorphic speech datasets including TI46-Alpha~\cite{nist-ti46-1991}, TI46-Digits~\cite{nist-ti46-1991}, and N-TIDIGITS~\cite{anumula2018feature}, the proposed ScSr-SNNs can boost performance by up to $2.55\%$ compared with other types of RSNNs trained by state-of-the-art BP methods.

\section{Background}
\subsection{Spiking Neuron Model}
SNNs employ a more biologically plausible spiking neuron model than ANNs. In this work, the leaky integrate-and-fire (LIF) neuron model~\cite{gerstner2002spiking} are adopted. 

First, we notate the input spike train from pre-synaptic neuron $j$ to post-synaptic neuron $i$ as  
\begin{equation}
\label{eq:neuron_input}
s_{j}(t)=\sum_{t_j^{(f)}}\delta(t-t_j^{(f)}),
\end{equation}
where $t_j^{(f)}$ denotes a firing time of presynaptic neuron $j$. 

Then, the incoming spikes are converted into an (unweighted) postsynaptic current (PSC) $a_j(t)$ through a synaptic model. The first order synaptic model~\cite{gerstner2002spiking} is considered and defined as
\begin{equation}
    \label{eq:first-order}
    \tau_s \frac{\alpha_j(t)}{dt} = -\alpha_j(t) + s_{j}(t),
\end{equation}
where $\tau_s$ is the synaptic time constant.

The neuronal membrane voltage $u_i(t)$ of post-synaptic neuron $i$ at time $t$ is given by
\begin{equation}
    \label{eq:lif}
    \tau_m \frac{du_i(t)}{dt} = -u_i(t) + R~\sum_j w_{ij}a_j (t), 
\end{equation}
where $R$ and $\tau_m$ are the effective leaky resistance and time constant of the membrane, $w_{ij}$ is the synaptic weight from neuron $j$ to neuron $i$, $a_j(t)$ is the (unweighted) postsynaptic potential (PSC) induced by the spikes from neuron $j$.

Considering the discrete time steps simulation, we use the fixed-step first-order Euler method to discretize (\ref{eq:lif}) to
\begin{equation}
    \label{eq:discrete_lif}
    u_i[t] = \theta_m u_i[t-1] + \sum_j w_{ij}a_j [t],
\end{equation}
where $\theta_m = 1 - \frac{1}{\tau_m}$. The ratio of R and $\tau_m$ is absorbed into the synaptic weight. Moreover, the firing output of the neuron $i$ is expressed as
\begin{equation}
    \label{eq:fire_function}
    s_i[t] = H\left(u_i[t] - V_{th}\right),
\end{equation}
where $V_{th}$ is the firing threshold and $H(\cdot)$ is the Heaviside step function.

\subsection{SNNs Forward Pass}
\begin{figure}[ht]
\begin{center}
\includegraphics[width=0.4\textwidth]{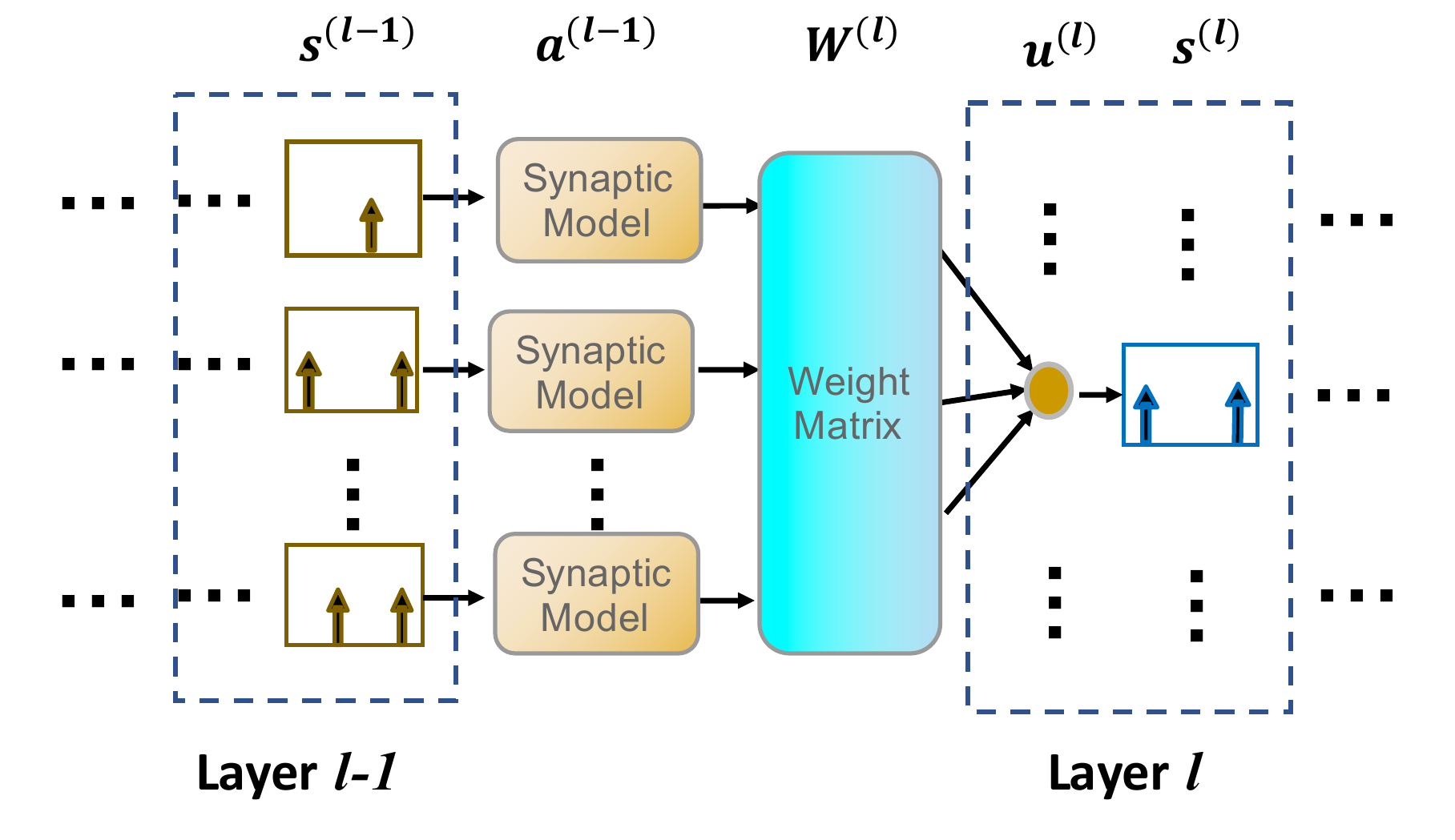}
\end{center}
\caption{Forward evaluation pass of SNNs.}
\label{fig:forward_pass}
\end{figure}

As shown in Fig \ref{fig:forward_pass}, we describe a feedforward neural network. Consider a layer $l$ with $N_l$ neurons, we denote the presynaptic weights by $\bm{W}^{(l)}=\left[\bm{w}^{(l)}_1, \cdots, \bm{w}^{(l)}_{N_{l}}\right]^T$, where $\bm{w}^{(l)}_{i}$ is a column vector of weights from all the neurons in layer $l-1$ to the neuron $i$ of layer $l$. 
In addition, we also denote PSCs from neurons in layer $l-1$  by $\bm{a}^{(l-1)}[t]=\left[a^{(l-1)}_{1}[t], \cdots, a^{(l-1)}_{N_{l-1}}[t]\right]^T$, output spike trains of the $l-1$ layer by $\bm{s}^{(l-1)}[t]=\left[s^{(l-1)}_{1}[t], \cdots, s^{(l-1)}_{N_{l-1}}[t]\right]^T$, membrane potentials of the $l$ layer neurons by $\bm{u}^{(l)}[t]=\left[u^{(l)}_{1}[t], \cdots, u^{(l)}_{N_{l}}[t]\right]^T$, where variables associated with the layer $l$ have $l$ as the superscript. Therefore, the network forward pass can be described as
\begin{equation}
    \label{eq:forward_pass}
    \begin{split}
        & \bm{a}^{(l-1)}[t] =  (1 - \frac{1}{\tau_s}) \bm{a}^{(l-1)}[t-1] + \bm{s}^{(l-1)}[t], \\ 
        & \bm{u}^{(l)}[t] =  \theta_m\bm{u}^{(l)}[t-1] + \bm{W}^{(l)}\bm{a}^{(l-1)}[t], \\
        & \bm{s}^{(l)}[t] = H\left(\bm{u}^{(l)}[t] - V_{th}\right),
    \end{split}
\end{equation}

In the forward pass, the spike trains $\bm{s}^{(l-1)}[t]$ of the $l-1$ layer generate the (unweighted) PSCs $\bm{a}^{(l-1)}[t]$ according to the synaptic model. Then, $\bm{a}^{(l-1)}[t]$ are multiplied the synaptic weights and passed onto the neurons of layer $l$. The integrated PSCs alter the membrane potentials and trigger the output spikes of the layer $l$ neurons when the membrane potentials exceed the threshold.

\section{Methods}
In this section, we first introduce the two components in ScSr-SNNs, the self-recurrent connections and the skip-connections. We illustrate that self-recurrent layers can realize recurrent behaviors similar to one recurrent layer while self-recurrent connections also implement extra local memory. Then, we introduce the skip-connected structure which can benefit neuronal dynamics and correlations. Finally, we demonstrate the BIP method which joint the BP method and IP method to further boost the performance.

\subsection{Self-recurrent Architecture} \label{sec:self_recurrent}
The idea of connecting neuron back to itself for ANNs is introduced in \cite{li2018independently}. It proposed the independently RNN (IndRNN) which is a deep ANN with independently recurrent connections in each layer. It is proved that Multiple IndRNNs can be stacked to construct a deep RNN. In this work, we apply a similar idea to SNNs. However, applying self-recurrence to SNNs not only constructs deep RSNNs but also implements local memory.

For simplicity, we adopt the method in~\cite{huh2018gradient} to replace the threshold and synaptic model with a gate function g(v). Thus, the PSC is defined as
\begin{equation}
    \bm{a}^{(l)}[t] = g\left(\bm{u}^{(l)}[t]\right),
\end{equation}

In a single layer, the computation of each neuron is independent while neurons can be correlated through multiple layers. Neurons in the same self-recurrent layer only recurrently connected to themselves and can be described as
\begin{equation}
    \label{eq:self_recurrent_layer}
    \begin{split}
        \bm{u}^{(l)}[t] & = \theta_m \bm{u}^{(l)}[t-1] \\
        & + \bm{W}^{(l)}\bm{a}^{(l-1)}[t] + \bm{W}_s^{(l)}g\left(\bm{u}^{(l)}[t-1]\right), 
    \end{split}
\end{equation}
where $\bm{W}_s^{(l)}=\left[w_{s, 1}^{(l)}, \cdots, w_{s, N_l}^{(l)}\right]$
is a vector weight matrix of self-recurrent connections in layer $l$ and $w_{s, i}^{(l)}$ is the weight of neuron $i$'s self-recurrent connection in layer $l$.

Two self-recurrent layers can work similarly to the fully connected recurrent layer. To illustrate this, we approximately derive the behavior of two self-recurrent layers and compare it with a fully connected recurrent layer.  

\begin{figure}[ht]
\begin{center}
\includegraphics[width=0.25\textwidth]{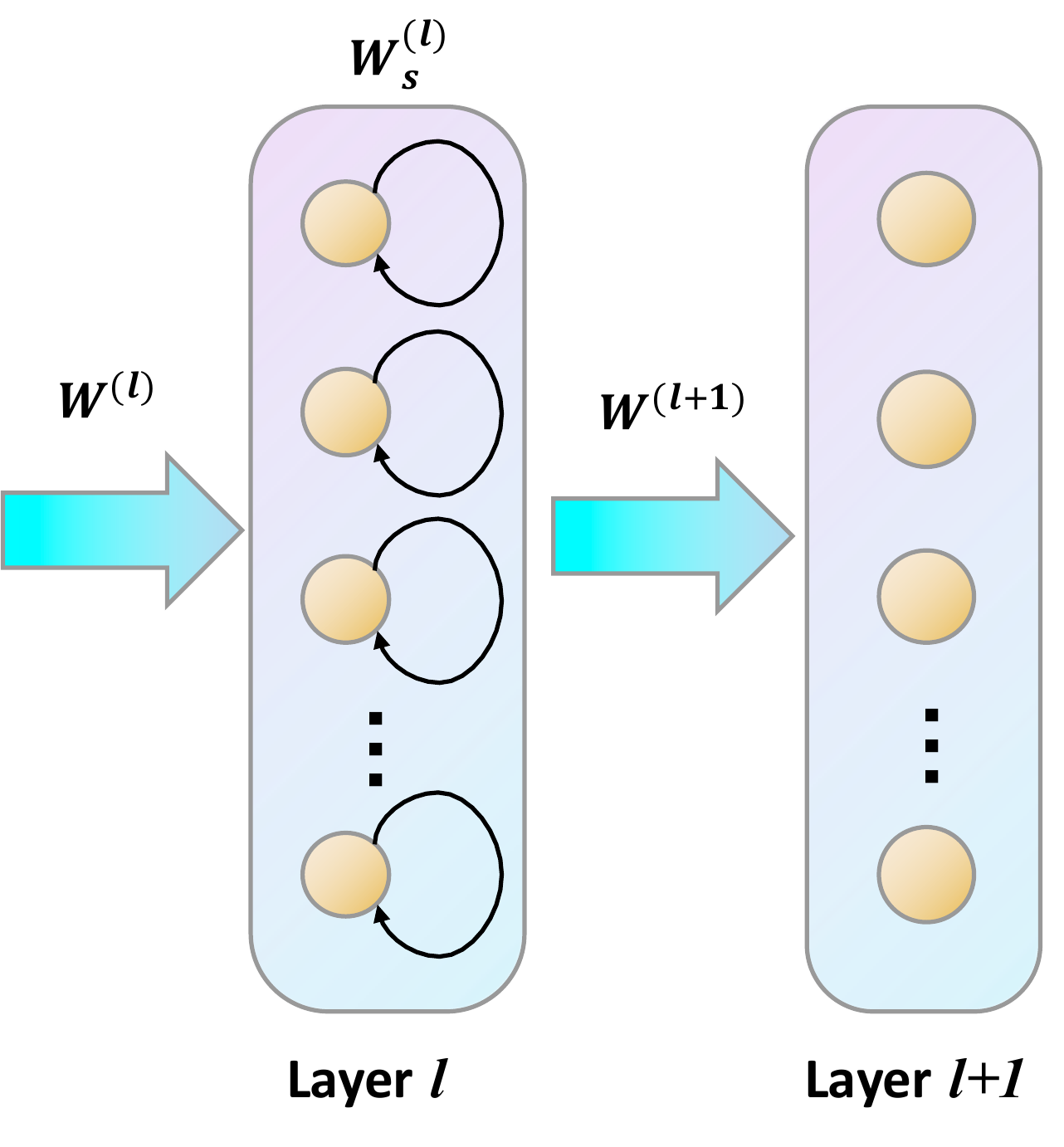}
\end{center}
\caption{Self-recurrent pass of SNNs.}
\label{fig:self_recurrent}
\end{figure}

Fig \ref{fig:self_recurrent} presents the network with layer $l$ and layer $l+1$ while layer $l$ has self-recurrent connections. In this case, the behavior can be expressed as
\begin{equation}
    \label{eq:two_layers_self_recurrent}
    \begin{split}
        \bm{u}^{(l)}[t] & = \theta_m \bm{u}^{(l)}[t-1] \\
        & + \bm{W}^{(l)}\bm{a}^{(l-1)}[t] + \bm{W}_s^{(l)}\bm{a}^{(l)}[t-1], \\
        \bm{u}^{(l+1)}[t] & = \theta_m \bm{u}^{(l+1)}[t-1] + \bm{W}^{(l+1)}g\left(\bm{u}^{(l)}[t]\right). 
    \end{split}
\end{equation}

More aggressively, we consider the g(v) as a linear function $g(v)=h\cdot v$. Then, the (\ref{eq:two_layers_self_recurrent}) can be combined as
\begin{equation}
    \label{eq:two_layers_self_recurrent_combined}
    \begin{split}
        \bm{u}^{(l+1)}[t] & = \theta_m \bm{u}^{(l+1)}[t-1] + \bm{W}^{(l)}\bm{a}^{(l-1)}[t] \\
         & + \bm{W}_1^{(l)}\bm{a}^{(l+1)}[t-1] + \bm{W}_2^{(l)}\bm{a}^{(l+1)}[t-2], \\
        \bm{W}_1^{(l)} & = \bm{W}^{(l+1)}(\frac{\theta_m}{h}+\bm{W}_s^{(l)})(\bm{W}^{(l+1)})^{-1} \\
        \bm{W}_2^{(l)} & = -\theta_m\bm{W}_1^{(l)}.
    \end{split}
\end{equation}
The behavior of two layers network illustrated in (\ref{eq:two_layers_self_recurrent_combined}) is similar to the single recurrent layer which is described as
\begin{equation}
\label{eq:rnn}
       \bm{u}^{(l)}[t] = \theta_m \bm{u}^{(l)}[t-1] + \bm{W}^{(l)}\bm{a}^{(l-1)}[t] + \bm{W}_r^{(l)}\bm{a}^{(l)}[t-1],
\end{equation}
where $\bm{W}_r^{(l)}$ is random recurrent connection weights matrix.

Compared to the (\ref{eq:rnn}), two layers self-recurrent structure can be viewed as a single recurrent layer with two groups of recurrent connections which have a delay of $1$ and $2$ timesteps. Note that (\ref{eq:two_layers_self_recurrent_combined}) also indicates the constraints of two layers self-recurrent structure that: 1. the recurrent weights $\bm{W}_1^{(l)}$ and $\bm{W}_2^{(l)}$ should be diagonalizable; 2. $\bm{W}_2^{(l)}$ is the negative of $\bm{W}_1^{(l)}$ with the scalar $\theta_m$.

Moreover, in the spiking neuron, a neuron will lose previous information after the it fires and resets the membrane voltage. As shown in (\ref{fig:self_recurrent}), the self-recurrent connections passing the output signals back to the original neurons can be view as a compensation to the information loss for the firing-and-resetting mechanism. Therefore, the local memory of the neuron is applied.

To sum it up, a two layers self-recurrent structure behave similarly to a traditional recurrent layer. However, compared to the traditional random connected RSNNs, ScSr-SNNs brings several benefits from the self-recurrent connections:

\textbullet\ ScSr-SNNs has a simpler but more structural architecture than the randomly generated RSNNs.

\textbullet\ The error gradients can be more straightforwardly calculated in ScSr-SNNs due to the mostly feedforward nature of the network. Thus, the straightforward calculation also cost less computational resources.

\textbullet\ The self-connection implements local memory by mitigating the memory loss due to the firing and resetting mechanism.

\textbullet\ From experimental results, the ScSr-SNNs shows better results than the same size RSNNs.

\subsection{Skip Connections}
\begin{figure}[ht]
\begin{center}
\includegraphics[width=0.35\textwidth]{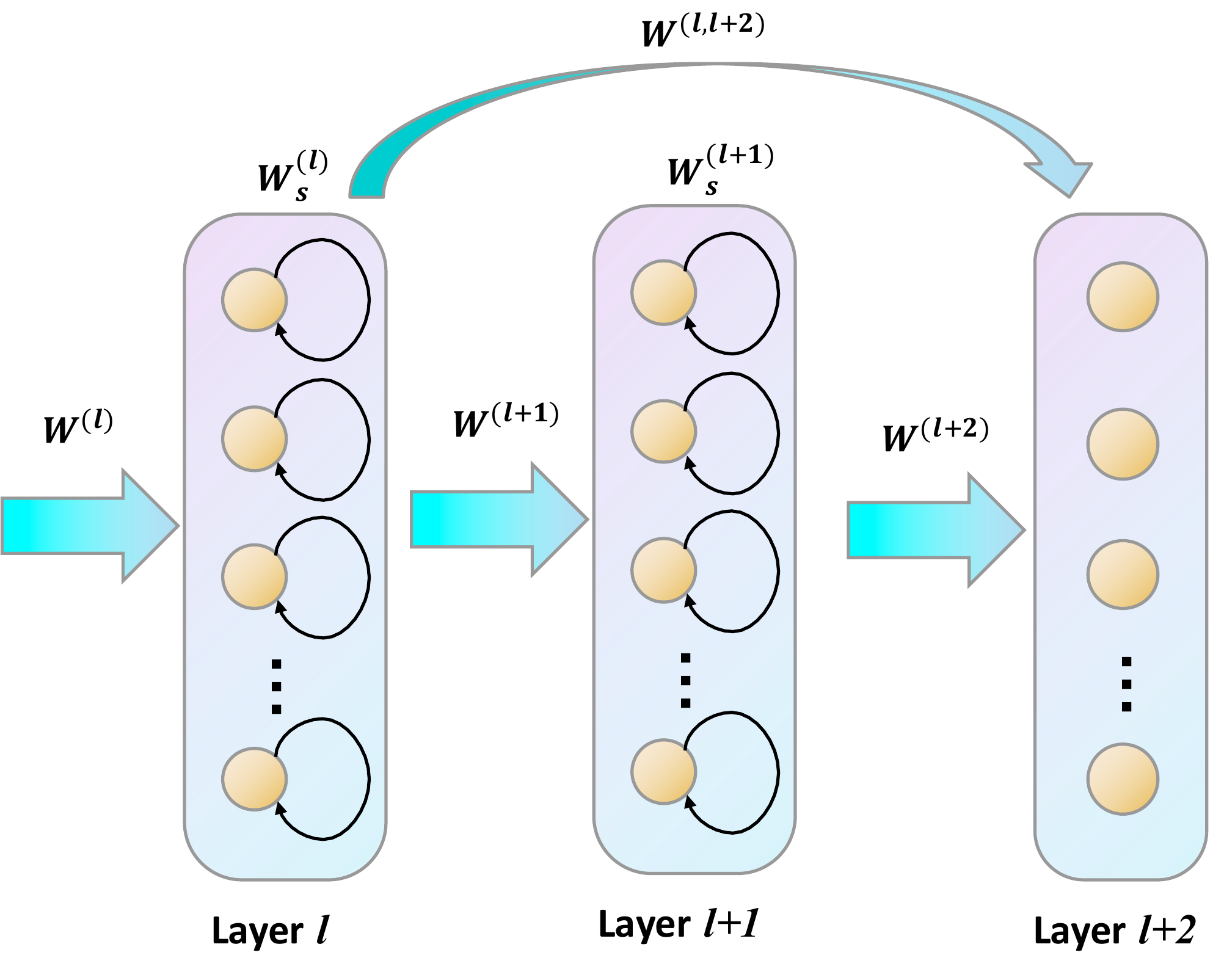}
\end{center}
\caption{Skip-connected pass of SNNs.}
\label{fig:skip_connections}
\end{figure}
As shown in Fig \ref{fig:skip_connections}, we adopt the skip-connections which directly connect two non-adjacent layers to enrich the network dynamics. The skip-connections benefit the network for three reasons. First, as introduced, two layers with self-recurrent connections can work similarly to a recurrent layer. Therefore, in Fig \ref{fig:skip_connections}, we can view layer $l$ and $l+1$ as a recurrent layer while the $l+1$ and $l+2$ layers also form an equivalent recurrent layer. With the additional skip-connections from layer $l$ to $l+2$, these two layers can also be treated as an approximate recurrent layer. The skip-connections bring more possibilities and dynamics to form different structures. Second, the skip-connections from layer $l$ to layer $l+2$, as well as the self-recurrent connections, also build the three layers into a large recurrent structure which further enriches the network dynamics.  Furthermore, the skip-connections can also be trained by BP method which lead to additional tunability for the network.

\subsection{Backpropagation for self-recurrent SNNs}
In this paper, we adopt the TSSL-BP method proposed in~\cite{zhang2020temporal} to train the ScSr-SNNs. TSSL-BP is a BPTT based backpropagation method for SNNs. It captures the error backpropagation across two types of inter-neuron and intra-neuron dependencies and leads to state-of-the-art performance with extremely low latency. In the original paper, the TSSL-BP is proposed for feedforward SNNs. In this paper, we extend its applicability to the self-recurrent connections and skip-connections. 

During the training, we denote the desired and the actual spike trains in the output layer by $\bm{d} = \left[\bm{d}[t_0], \cdots, \bm{d}[t_{N_t}]\right]$ and $\bm{s} = \left[\bm{s}[t_0], \cdots, \bm{s}[t_{N_t}]\right]$ where $N_t$ is the number of the considered time steps, $\bm{d}[t]$ and $\bm{s}[t]$ the desired and actual firing events for all output neurons at time $t$, respectively. Note that in this work, the output layer has neither self-recurrent connections nor skip-connections.

The loss function $L$ is defined by the square error for each output neuron at each time step:
\begin{equation}
\label{eq:loss_function}
        L = \sum_{k=0}^{N_t}E[t_k] = \sum_{k=0}^{N_t}\frac{1}{2}((\epsilon * \bm{d})[t_k] - (\epsilon * \bm{s})[t_k])^2,
\end{equation}
where $\epsilon(\cdot)$ is a kernel function which measures the so-called Van Rossum distance between the actual spike train and desired spike train. 

We consider the backpropagation in three cases: 1. feedforward layer, 2. self-recurrent layer, 3. self-recurrent layer with skip-connection to a post layer.

\textbf{[Case 1: feedforward layer]}  As shown in Fig \ref{fig:forward_pass}, when the layer $l$ is feedforward layer, it could be the output layer or a hidden layer.

For the neurons in layer $l$, (\ref{eq:forward_pass}) reveals that the values of $\bm{u}^{(l)}$ at time $t$ have contribution to all future fires and losses. Therefore, the error gradient with respect to the presynaptic weights matrix $\bm{W}^{(l)}$ can be described as
\begin{equation}
    \label{eq:d_l_d_w}
    \begin{split}
        & \frac{\partial L}{\partial \bm{W}^{(l)}}  = \sum_{k=0}^{N_t}\frac{\partial E[t_k]}{\partial\bm{W}^{(l)}} = \sum_{k=0}^{N_t}\sum_{m=0}^{k}\frac{\partial E[t_k]}{\partial \bm{u}^{(l)}[t_m]} \frac{\partial \bm{u}^{(l)}[t_m]}{\partial\bm{W}^{(l)}} \\
        & = \sum_{m=0}^{N_t}\bm{a}^{(l-1)}[t_m] \sum_{k=m}^{N_t}\frac{\partial E[t_k]}{\partial \bm{u}^{(l)}[t_m]}.  
    \end{split}
\end{equation}

We use $\delta$ to denote the back propagated error at time $t_m$: 
\begin{equation}
    \label{eq:delta_output_layer}
    \bm{\delta}^{(l)}[t_m]= \sum_{k=m}^{N_t}\frac{\partial E[t_k]}{\partial \bm{u}^{(l)}[t_m]} = \sum_{k=m}^{N_t}\frac{\partial E[t_k]}{\partial \bm{a}^{(l)}[t_k]}\frac{\partial \bm{a}^{(l)}[t_k]}{\partial \bm{u}^{(l)}[t_m]}.
\end{equation}

If $l$ is the output layer, from (\ref{eq:loss_function}), the first term of (\ref{eq:delta_output_layer}) is given by
\begin{equation}
\frac{\partial E[t_k]}{\partial \bm{a}^{(l)}[t_k]} = (\epsilon * \bm{d})[t_k] - (\epsilon * \bm{s})[t_k].
\end{equation}

If $l$ is a hidden layer, $\bm{\delta}^{(l)}[t_m]$ is derived from: 
\begin{equation}
    \label{eq:case1_delta}
    \begin{split}
        & \bm{\delta}^{(l)}[t_m] = \sum_{j=m}^{N_t}\sum_{k=m}^{j}\frac{\partial \bm{a}^{(l)}[t_k]}{\partial \bm{u}^{(l)}[t_m]}\left(\frac{\partial \bm{u}^{(l+1)}[t_k]}{\partial \bm{a}^{(l)}[t_k]}\frac{\partial E[t_j]}{\partial \bm{u}^{(l+1)}[t_k]} \right) \\
        & = \sum_{k=m}^{N_t}\frac{\partial \bm{a}^{(l)}[t_k]}{\partial \bm{u}^{(l)}[t_m]}\sum_{j=k}^{N_t}\bm{W}^{(l+1)}\frac{\partial E[t_j]}{\partial \bm{u}^{(l+1)}[t_k]} \\ 
        & = (\bm{W}^{(l+1)})^T\sum_{k=m}^{N_t}\frac{\partial \bm{a}^{(l)}[t_k]}{\partial \bm{u}^{(l)}[t_m]}\bm{\delta}^{(l+1)}[t_k].
    \end{split}
\end{equation}

(\ref{eq:case1_delta}) demonstrate that membrane potentials $\bm{u}^{(l)}$ of the neurons in layer $l$ influence their (unweighted) PSCs $\bm{a}^{(l)}$ through fired spikes, and $\bm{a}^{(l)}$ further affect the membrane potentials $\bm{u}^{(l+1)}$ in the next layer. 

The calculation of the term $\frac{\partial \bm{a}^{(l)}[t_k]}{\partial \bm{u}^{(l)}[t_m]}$ is one of the key contributions of~\cite{zhang2020temporal}. We do not repeat the derivatives here but treat it as a known term in the rest of this paper.

\textbf{[Case 2: self-recurrent layer]}
The structure of the self-recurrent layer is shown in Fig \ref{fig:self_recurrent}. Similar to the feedforward case, the weights of the incoming synapses of the pre layer can be calculated according to (\ref{eq:d_l_d_w}). Based on (\ref{eq:self_recurrent_layer}) the error gradient with respect to the weights of self recurrent layer can be expressed as:
\begin{equation}
    \label{eq:d_l_d_ws}
        \frac{\partial L}{\partial \bm{W}_s^{(l)}}  = \sum_{m=1}^{N_t}\bm{a}^{(l)}[t_m-1]\bm{\delta}^{(l)}[t_m].  
\end{equation}

Compared to the feedforward case, the main difference comes from the derivation of $\bm{\delta}^{(l)}[t_m]$. Except the error signals from the post layer, we also need to take the errors backpropagated from self-recurrent connections into consideration. Thus, the error $\bm{\delta}^{(l)}[t_m]$ is calculated by:
\begin{equation}
    \label{eq:case2_delta}
    \begin{split}
        & \bm{\delta}^{(l)}[t_m] = \sum_{j=m}^{N_t}\sum_{k=m}^{j}\frac{\partial \bm{a}^{(l)}[t_k]}{\partial \bm{u}^{(l)}[t_m]}\left(\frac{\partial \bm{u}^{(l+1)}[t_k]}{\partial \bm{a}^{(l)}[t_k]}\frac{\partial E[t_j]}{\partial \bm{u}^{(l+1)}[t_k]} \right) \\
        & + \sum_{j=m+1}^{N_t}\sum_{k=m}^{j}\frac{\partial \bm{a}^{(l)}[t_k]}{\partial \bm{u}^{(l)}[t_m]}\left(\frac{\partial \bm{u}^{(l)}[t_k+1]}{\partial \bm{a}^{(l)}[t_k]}\frac{\partial E[t_j]}{\partial \bm{u}^{(l)}[t_k+1]} \right) \\
        & = (\bm{W}^{(l+1)})^T\sum_{k=m}^{N_t}\frac{\partial \bm{a}^{(l)}[t_k]}{\partial \bm{u}^{(l)}[t_m]}\bm{\delta}^{(l+1)}[t_k] \\
        & + (\bm{W_s}^{(l)})^T\sum_{k=m}^{N_t}\frac{\partial \bm{a}^{(l)}[t_k]}{\partial \bm{u}^{(l)}[t_m]}\bm{\delta}^{(l)}[t_k+1].
    \end{split}
\end{equation}

In (\ref{eq:case2_delta}), the first term is the same as (\ref{eq:case1_delta}) which represent the error backpropagated from post layer. The second term is caused by self-recurrent connections. It reveals that membrane potentials $\bm{u}^{(l)}$ of the neurons in layer $l$ influence their (unweighted) PSCs $\bm{a}^{(l)}$ through fired spikes, and $\bm{a}^{(l)}$ further affect the membrane potentials of $\bm{u}^{(l)}$ at the next timestep. This part of error is backpropapageted  through time via the self-recurrent connections.

\textbf{[Case 3: self-recurrent layer with skip-connections to a post layer]}
As shown in Fig \ref{fig:skip_connections}, we suppose layer $l$ has self-recurrent connections. In addition, it also connects to layer $l+1$ the same as the feedforward layer and layer $l+2$ through the skip-connections. The changes of BP method still come from the derivation of $\bm{\delta}^{(l)}[t_m]$. In this case, the output (unweighted) PSCs $\bm{a}^{(l)}$ of layer $l$ further directly affects the membrane potentials of $\bm{u}^{(l+2)}$ at layer $l+2$.

Similar to the derivation of (\ref{eq:case2_delta}), an additional term should be added when the skip-connections are taken into account. Therefore, the $\bm{\delta}^{(l)}[t_m]$ can be derived as
\begin{equation}
    \label{eq:case3_delta}
    \begin{split}
        & \bm{\delta}^{(l)}[t_m] = (\bm{W}^{(l+1)})^T\sum_{k=m}^{N_t}\frac{\partial \bm{a}^{(l)}[t_k]}{\partial \bm{u}^{(l)}[t_m]}\bm{\delta}^{(l+1)}[t_k] \\
        & + (\bm{W_s}^{(l)})^T\sum_{k=m}^{N_t}\frac{\partial \bm{a}^{(l)}[t_k]}{\partial \bm{u}^{(l)}[t_m]}\bm{\delta}^{(l)}[t_k+1] \\
        & + (\bm{W}^{(l,l+2)})^T\sum_{k=m}^{N_t}\frac{\partial \bm{a}^{(l)}[t_k]}{\partial \bm{u}^{(l)}[t_m]}\bm{\delta}^{(l+2)}[t_k],
    \end{split}
\end{equation}
where $\bm{W}^{(l,l+2)}$ represents the skip-connections from layer $l$ to layer $l+2$.

\subsection{Backpropagation based Intrinsic Plasticity}
Intrinsic plasticity (IP) is a widely used self-adaptive mechanism that maintains homeostasis and shapes the dynamics of neural circuits. In short, IP is an inner neuronal mechanism that adjusts the neuron's activity. IP has been observed in various biological neurons~\cite{marder1996memory,baddeley1997responses,desai1999plasticity}. However, there's still no conclusion about how the IP exactly works. 

The most straight forward idea is that IP boosts neuron's activity if the neuron rarely fires and depresses the neuron if it fires too frequently. In SNNs, there are mainly two ways to apply the IP method. First, the most famous way is called the dynamic threshold. It heuristically increases the neuronal firing threshold when the neuron spikes and exponentially decays the firing threshold during the rest of the time~\cite{lazar2007fading, bellec2018long,  li2018computational}. On the other hand, \cite{triesch2005gradient} first proposed a mathematical IP rule on ANNs to adjust the neuron so that its Kullback–Leibler (KL) divergence from an exponential distribution to the actual output firing rate distribution is minimized. It means that the rule adjusts the neuron to generate responses following the exponential distribution. After that several works tried to apply this idea to SNNs by adjusting the neuronal parameters to optimize the neuron's output firing rate distribution~\cite{schrauwen2008improving, li2013spike, zhang2019information}.

In \cite{zhang2019information}, the IP method trains the time constant to minimize the loss which is defined by the KL divergence between the desired and actual firing rate distributions. Similarly, in this work, we also train the time constant of the neuron to control the neuron activity. However, the original IP method in \cite{zhang2019information} targets neuron's firing rate distribution. It may contradict the objective of the BP rule that tries to minimize the output loss. Therefore, in the work, we propose a backpropagated intrinsic plasticity (BIP) method to joint both the IP and BP methods for the same goal that is the output loss. 

Therefore, according to (\ref{eq:forward_pass}), the neuronal parameter $\theta_m$ of each neuron are trained with the backpropagated errors. We rewrite the forward pass with self-recurrent connections as
\begin{equation}
    \label{eq:self_recurrent_layer_IP}
        \bm{u}^{(l)}[t] = \bm{\theta}^{(l)} \bm{u}^{(l)}[t-1]  + \bm{W}^{(l)}\bm{a}^{(l-1)}[t] + \bm{W}_s^{(l)}\bm{a}^{(l)}[t-1], 
\end{equation}
where $\bm{\theta}^{(l)}=\left[\theta^{(l)}_{1}, \cdots, \theta^{(l)}_{N_{l}}\right]^T$ is the vector of time constant for all neurons in layer $l$. 

Similar to (\ref{eq:d_l_d_w}), the $\bm{\theta}^{(l)}$ is updated based on the backpropagated error $\bm{\delta}^{(l)}$. The error gradient can be described as 
\begin{equation}
    \label{eq:d_l_d_w_skip_connections}
    \begin{split}
        \frac{\partial L}{\partial \bm{\theta}^{(l)}}  = \sum_{m=1}^{N_t}\bm{u}^{(l)}[t_m-1]\bm{\delta}^{(l)}[t_m].  
    \end{split}
\end{equation}

In this paper, we proposed the ScSr-SNNs which is a new structure of RSNNs with self-recurrent connections as well as the skip-connections. In addition, we extend the feedforward SNNs BP method TSSL-BP to be eligible to train ScSr-SNNs. Furthermore, we use the BIP method which combines the IP and BP to further boost the training performance.

\section{Experiments and Results}
\subsection{Experimental Settings}
In this section, we test the proposed ScSr-SNNs with BIP method on three datasets, the speech dataset TI46-Alpha~\cite{nist-ti46-1991}, TI46-Digits~\cite{nist-ti46-1991} and Neuromorphic speech dataset N-TIDIGITS~\cite{anumula2018feature}. We compare ScSr-SNNs with several state-of-the-art results with the same or similar network sizes including feedforward SNNs, RSNNs, Liquid State Machine(LSM), and ANNs. 

All reported experiments are conducted on an NVIDIA Titan XP GPU. The implementation of ScSr-SNNs is on the Pytorch framework~\cite{NEURIPS2019_9015}. The simulation step size is set to $1$ ms. 
The parameters like thresholds and learning rates are empirically tuned. Table \ref{tab:parameters_settings} lists the typical constant values adopted in our experiments for each dataset. No axon and synaptic delay or refractory period is adopted in the feedforward pass whereas the self-recurrent connections has $1$ ms delay. No dropout or normalization is applied. Adam~\cite{kingma2014adam} is adopted as the optimizer. 

\begin{table}[ht]
\centering
\begin{small}
\begin{tabular}{l|c|c|c}
\toprule
\textbf{Parameter} & \textbf{TI46-Alpha} & \textbf{TI46-Digit} & \textbf{N-TIDIGITS}\\
\midrule
$\tau_{m}\textsuperscript{a}$    & 16 ms & 16 ms & 64 ms \\
$\tau_{s}$    & 8ms & 8 ms & 8 ms \\
learning rate & 0.005 & 0.005 & 0.002 \\
Threshold \textit{$V_{th}$} & 1 mV &  1 mV   & 1 mV \\
Batch Size  & 50  & 50   &  50  \\
Time steps  & 100  & 100   &  300   \\
Epochs & 400  &  400  &  400  \\
\bottomrule
\multicolumn{4}{l}{\footnotesize{\textsuperscript{a}For BIP method, the $\tau_m$ here means the initial value.}}
\end{tabular}
\end{small}
\caption{Parameters settings.}
\label{tab:parameters_settings}
\end{table}

For practical issues of implementing the BP method such as desired output selection, warm-up mechanism, the boundary of derivatives, we follow the solutions in \cite{zhang2020temporal}.

We will share our Pytorch implementation on GitHub. We expect this work will contribute to the exploration of different SNNs structures and learning methods.

\subsection{Datasets and Preprocessing}
The speech recognition task usually contains highly related context through time. Therefore, recurrent networks are thought to be the proper one for these tasks. In this paper, we test the proposed ScSr-SNNs on the speech datasets TI46-Alpha~\cite{nist-ti46-1991}, TI46-Digits~\cite{nist-ti46-1991} and Neuromorphic speech dataset N-TIDIGITS~\cite{anumula2018feature}.

\textbullet\ TI46-Alpha: TI46-Alpha is the full alphabets subset of the TI46 Speech corpus~\cite{nist-ti46-1991} and contains spoken English alphabets from $16$ speakers. There are  4,142 and 6,628 spoken English examples in 26 classes for training and testing, respectively. The continuous temporal speech waveforms are preprocessed by Lyon's ear model~\cite{lyon1982computational}. Each speech is encoded into 78 channels with real-value intensity. In this case, we directly apply the converted real-value inputs to the input layer. In Lyon's ear model, the sample rate is chosen such that each channel in a sample has $100$ values which represent the signals in $100$ timesteps.

\textbullet\ TI46-Digit: TI46-Digits is the full digits subset  of the TI46 Speech corpus~\cite{nist-ti46-1991}. It contains  1,594 training examples and 2,542 testing examples of 10 utterances for each of digits "0" to "9" spoken by 16 different speakers. The same preprocessing used for TI46-Alpha is adopted.

\textbullet\ N-TIDIGITS: The N-Tidigits~\cite{anumula2018feature} is the neuromorphic version of the well-known speech dataset Tidigits~\cite{leonard1993tidigits}, and consists of recorded spike responses of a 64-channel CochleaAMS1b sensor in response to the original audio waveforms. The dataset includes both single digits and connected digit sequences with a vocabulary consisting of 11 digits including “oh,” “zero” and the digits “1” to “9”. There are 55 male and 56 female speakers with 2,475 single digit examples for training and the same number of examples for testing. In the original dataset, each sample lasts about $0.9s$. In our experiments, we reduce the time resolution to speed up the simulation. Therefore, the preprocessed samples only have about $300$ time steps. We determine that a channel has a spike at a certain time step of the preprocessed sample if there's at least one spike among the corresponding several time steps of the original sample.

\subsection{Experimental Results} \label{sec:experimental_results}
In this section, we demonstrate the performance of ScSr-SNNs with the BIP method. For each result, the mean and standard deviation (stddev) reported are obtained by repeating the experiments five times. Moreover, we also separately apply the proposed self-recurrent structure, skip-connected structure, and BIP method to the networks to reveal their individual effects of boosting performance. For all the results reported in this paper, we apply the SNNs with 3 hidden layers. Without specifying, we follow Table \ref{tb:structures} to name the structures and methods. 

\begin{table}[ht]
\centering
\begin{small}
\begin{tabular}{l|c}
\toprule
\textbf{Name} & \textbf{Structures} \\
\midrule
 Sr-SNNs  &  three self-recurrent layers\\
 Sr-SNNs-BIP & Sr-SNNs with BIP \\
 SrSc-SNNs & Sr-SNNs, skip-connections from layer 1 to 3 \\
 SrSc-SNNs-BIP & SrSc-SNNs with BIP\\
\bottomrule
\end{tabular}
\end{small}
\caption{Structures.}
\label{tb:structures}
\end{table}

\textbullet\ TI46-Alpha: As shown in Table \ref{tb:ti46-alpha}, we compare the proposed structures with several existing results. In \cite{wijesinghe2019analysis}, the state vector of the reservoir is used to train the single readout layer using BP. Its result shows that only training the single readout layer of a  recurrent  LSM is inadequate for this challenging task, demonstrating the necessity of a more structural SNN. \cite{zhang2019spike}  demonstrates the state-of-the-art performance of TI46-Digits. It has one recurrent layer trained with the ST-RSBP method. However, both recurrent networks in \cite{wijesinghe2019analysis} and \cite{zhang2019spike} are randomly generated. We show that, with the proposed SrSc-SNNs and BIP method, we can achieve a performance of $95.90\%$ with with a mean of $95.61\%$ and a standard deviation of $0.18\%$ which is $2.55\%$ better than the best-reported result. In the table, we also demonstrate that the three proposed methods, the self-recurrent connections, skip-connections, and BIP method, can all improve performance when they are applied separately. In addition, with the same training rule TSSL-BP, our proposed methods can boost $2.76\%$ performance on the same network size compared to \cite{zhang2020temporal}.

\begin{table}[ht]
\centering
\begin{small}
\begin{tabular}{ll}
\toprule
Structure     & Best  \\ \midrule
400-400~\cite{jin2018hybrid}   &   90.60\%       \\
LSM: R2000\textsuperscript{a}~\cite{wijesinghe2019analysis}  &   78\%    \\ 
400-R400-400~\cite{zhang2019spike}  &  93.35\%     \\
400-400-400~\cite{zhang2020temporal}   &  93.14\%   \\
Sr-SNNs: 400-400-400  &   94.51\%  \\
Sr-SNNs-BIP: 400-400-400   &   95.04\%  \\
SrSc-SNNs: 400-400-400  &   94.75\%  \\
SrSc-SNNs-BIP: 400-400-400   &  \textbf{95.90\%}   \\
\bottomrule
\multicolumn{2}{l}{\footnotesize{\textsuperscript{a} R represent recurrent layer.}} \\
\end{tabular}
\end{small}
\caption{Performances on TI46-Alpha}
\label{tb:ti46-alpha}
\end{table}

\textbullet\ TI46-Digit: For this dataset, we also compare the proposed methods with several latest results. The SpiLinC proposed in \cite{srinivasan2018spilinc} is an LSM with multiple reservoirs in parallel. Weights between input and reservoirs are trained using STDP. The excitatory neurons in the reservoir are tagged with the classes for which they spiked at the highest rate during training. The neurons with the same tag are grouped accordingly for inference. As shown in Tabel \ref{tb:ti46-digits}, with the proposed methods, We can achieve $99.69\%$ with a mean of $99.52\%$ and a standard deviation of $0.13\%$ which is $0.3\%$ better than the best-reported results in \cite{zhang2019spike} with only half of the network size. It proves that the proposed methods can improve the network dynamic when processing sequential signals and thus boost the performance.

\begin{table}[ht]
\centering
\begin{small}
\begin{tabular}{ll}
\toprule
Structures     & Best  \\ \midrule
LSM: R3200\cite{srinivasan2018spilinc}   &   86.66\%       \\
LSM: R5000\cite{wijesinghe2019analysis}    &   78\%    \\ 
200-R200-200\cite{zhang2019spike}   &  99.39\%   \\
100-100-100\cite{zhang2020temporal}    & 98.66\%   \\
Sr-SNNs: 100-100-100  &   99.17\%  \\
Sr-SNNs-IP: 100-100-100   &   99.31\%  \\
SrSc-SNNs: 100-100-100  &   99.35\%  \\
SrSc-SNNs-IP: 100-100-100  &  \textbf{99.69\%}  \\
\bottomrule
\end{tabular}
\end{small}
\caption{Performances on TI46-Digits}
\label{tb:ti46-digits}
\end{table}

\textbullet\ N-TIDIGITS: For N-TIDIGITS dataset, we not only compare the performance with previous work on RSNNs, but also with the well-known RNNs in non-spiking networks such as Gated Recurrent Unit (GRU) and long-short term memory (LSTM). The GRU and LSTM netowrks are trained by non-spiking BP method. As shown in Table \ref{tb:n-tidigits}, the proposed methods achieve $95.07\%$ with a mean of $94.79\%$ and a standard deviation of $0.22\%$ which is the best results compared to state-of-the-art performances. Note that although the neurons of the GRU and LSTM networks in Table \ref{tb:n-tidigits} are less than our proposed methods. The number of tunable parameters is nearly the same.

\begin{table}[ht]
\centering
\begin{small}
\begin{tabular}{ll}
\toprule
Structures    & Best  \\ \midrule
G200-G200-100\textsuperscript{a}~\cite{anumula2018feature}   &   89.69\%       \\
250L-250L\textsuperscript{b}~\cite{anumula2018feature}   &   90.90\%    \\ 
400-R400-400~\cite{zhang2019spike}  &  93.90\%   \\
400-400-400~\cite{zhang2020temporal}    & 89.85\%   \\
Sr-SNNs: 400-400-400  &   94.02\%  \\
Sr-SNNs-IP: 400-400-400  &   94.35\%  \\
SrSc-SNNs: 400-400-400  &   94.46\%  \\
SrSc-SNNs-IP: 400-400-400  &  \textbf{95.07\%}  \\
\bottomrule
\multicolumn{2}{l}{\footnotesize{\textsuperscript{a} G represents Gated Recurrent Unit (GRU) layer.}} \\
\multicolumn{2}{l}{\footnotesize{\textsuperscript{b} L represents long-short term memory (LSTM) layer.}}\\
\end{tabular}
\end{small}
\caption{Performances on N-TIDIGITS}
\label{tb:n-tidigits}
\end{table}

\section{Conclusion}
In this work, we propose the novel Skip-Connected Self-Recurrent SNNs (ScSr-SNNs) with backpropagated intrinsic plasticity (BIP). The self-recurrent connections no only implement the local memory by compensating the voltage drop because of the firing-and-resetting but also enhance network dynamics from the recurrence. In addition, the skip-connections also lead to more correlations between non-adjacent layers and thus further enrich dynamics. Moreover, the BIP together with BP method can improve the learning performance by concentrating on the output loss.

On the two speech datasets and a neuromorphic speech dataset, We demonstrate that the proposed ScSr-SNNs with the BIP method significantly outperformance the existing works including the spike train level BP method in \cite{zhang2019spike}. These great boosts also reveal the effectiveness of the proposed structures and methods. 

This work has been prototyped based on the widely adopted Pytorch framework and will be made available to the public.  We believe the proposed structures and BIP method will benefit the brain-inspired computing community from both a structural and algorithmic perspective.

\end{document}